\documentclass[conference,a4paper]{IEEEtran}
\IEEEoverridecommandlockouts
\usepackage{cite}
\usepackage{amsmath,amssymb,amsfonts}
\usepackage{algorithmic}
\usepackage{algorithm}
\usepackage{graphicx}
\usepackage{textcomp}
\usepackage{booktabs}
\usepackage{dirtytalk}
\usepackage{tabularx}
\usepackage{multirow}
\usepackage{makecell}
\usepackage{hyperref}
\usepackage{url}
\usepackage{xcolor}
\def\BibTeX{{\rm B\kern-.05em{\sc i\kern-.025em b}\kern-.08em
    T\kern-.1667em\lower.7ex\hbox{E}\kern-.125emX}}
\begin{document}

\title{NASiam: Efficient Representation Learning using Neural Architecture Search for Siamese Networks
\thanks{This work was performed using HPC resources from GENCI-IDRIS (Grant 20XX-AD011012644).}
}

\author{\IEEEauthorblockN{Alexandre Heuillet}
\IEEEauthorblockA{\textit{IBISC} \\
\textit{Université Paris-Saclay, Univ Evry}\\
Evry-Courcouronnes, France \\}
\IEEEauthorblockA{\textit{MRL}\\
\textit{Massachusetts Institute of Technology}\\
Cambridge, MA, USA \\
alexandre.heuillet@univ-evry.fr}
\and
\IEEEauthorblockN{Hedi Tabia}
\IEEEauthorblockA{\textit{IBISC} \\
\textit{Université Paris-Saclay, Univ Evry}\\
Evry-Courcouronnes, France \\
hedi.tabia@univ-evry.fr}
\and
\IEEEauthorblockN{Hichem Arioui}
\IEEEauthorblockA{\textit{IBISC} \\
\textit{Université Paris-Saclay, Univ Evry}\\
Evry-Courcouronnes, France \\
hichem.arioui@univ-evry.fr}
}

\maketitle

\begin{abstract}
Siamese networks are one of the most trending methods to achieve self-supervised visual representation learning (SSL). Since hand labeling is costly, SSL can play a crucial part by allowing deep learning to train on large unlabeled datasets. Meanwhile, Neural Architecture Search (NAS) is becoming increasingly important as a technique to discover novel deep learning architectures. However, early NAS methods based on reinforcement learning or evolutionary algorithms suffered from ludicrous computational and memory costs. In contrast, differentiable NAS, a gradient-based approach, has the advantage of being much more efficient and has thus retained most of the attention in the past few years. In this article, we present NASiam, a novel approach that uses for the first time differentiable NAS to improve the multilayer perceptron projector and predictor (encoder/predictor pair) architectures inside siamese-networks-based contrastive learning frameworks (e.g., SimCLR, SimSiam, and MoCo) while preserving the simplicity of previous baselines. We crafted a search space designed explicitly for multilayer perceptrons, inside which we explored several alternatives to the standard ReLU activation function. We show that these new architectures allow ResNet backbone convolutional models to learn strong representations efficiently. NASiam reaches competitive performance in both small-scale (i.e., CIFAR-10/CIFAR-100) and large-scale (i.e., ImageNet) image classification datasets while costing only a few GPU hours. We discuss the composition of the NAS-discovered architectures and emit hypotheses on why they manage to prevent collapsing behavior. Our code is available on \href{https://github.com/aheuillet/NASiam}{GitHub}.
\end{abstract}


\section{Introduction}

Deep Learning (DL) has experienced rapid growth in the past few years. Two DL subfields have received much attention: Unsupervised Representation Learning and Automated Deep Learning (AutoDL). 

Unsupervised representation learning aims to make DL models learn strong representations from unlabeled data. This is especially useful when considering that data labeling is often a costly and laborious human-made process. One of the most common approaches to unsupervised visual representation learning is siamese networks \cite{bromley1993signature}. Siamese networks consist of two weight-sharing branches (i.e., "twins") applied to two or more inputs. The output feature vectors of the two branches are compared to compute a loss (e.g., a \say{contrastive} loss). In the case of representation learning, the inputs are usually data augmentations of the same image, and the siamese networks seek to maximize the similarity between the output feature vectors of the two branches \cite{chen2020simple,chen2021progressive,chen2020improved}.  

On the other hand, AutoDL tries to remove the human factor from the DL pipeline. Architecture design is one part of this pipeline that has proven particularly relevant to automate. Most DL architectures are handcrafted and lack the certainty of an optimal solution \cite{chollet2017xception,szegedy2017inception,he2016deep}. Neural Architecture Search (NAS) aims to solve this issue by using a meta-learner to search for neural network architectures relevant to a given task (e.g., image classification, semantic segmentation, or object detection). NAS algorithms efficiently browse large search spaces that would prove challenging to navigate manually. The first NAS works used reinforcement learning \cite{zoph2017neural,zoph2018learning} or evolutionary methods \cite{chen2019renas,real2019regularized} but proved particularly inefficient, with thousands of GPU days needed to obtain a competitive architecture (e.g., 2000 GPU days for NASNet \cite{zoph2017neural}). Nowadays, most approaches use differentiable NAS \cite{chu2020fair,ye2022beta,dai2021fbnetv3} (i.e., gradient-based search process) as it requires far less computational resources and often yields better results.

This article leverages differentiable NAS to discover encoder (projector) and predictor architectures (i.e., Multilayer Perceptrons) that enable backbone Convolutional Neural Networks (CNNs) to efficiently learn strong representations from unlabeled data. To the extent of our knowledge, this is the first time that NAS has been applied to enhance the architecture of Siamese networks. Thus, we improved the performance of several siamese network frameworks such as SimSiam \cite{chen2021exploring}, SimCLR \cite{chen2020simple}, or MoCo \cite{chen2020improved}, with an encoder-predictor pair discovered by a meta-learner inspired by DARTS \cite{liu2019darts}, a popular differentiable NAS method. We dubbed our approach NASiam (\say{Neural Architecture Search for Siamese Networks}). We show that NASiam reaches competitive results on small-scale (CIFAR-10, CIFAR-100 \cite{krizhevsky2009learning}) and large-scale (ImageNet \cite{ILSVRC15}) datasets.
Thus, Section~\ref{sec:approach} highlights our main following contributions:
\begin{itemize}
    \item A novel way to design encoder/predictor pairs for siamese networks using differentiable neural architecture search.
    \item A novel search space specifically designed for the Multi-Layer Perceptron (MLP) heads of encoder/predictor pairs.
\end{itemize}
The rest of the article is structured as follows: Section~\ref{sec:related_work} features a short survey on related differentiable NAS and siamese networks works. In Section~\ref{sec:approach}, the proposed method is presented. Section~\ref{seq:experiments} presents the results of different image classification experiments and showcases a discussion on the composition of the discovered encoder/predictor pair architectures, and Section~\ref{sec:conclusion} brings a conclusion to our work while giving some insights about future work. 

\section{Related Work}
\label{sec:related_work}
This section briefly recalls related work in differentiable NAS, siamese networks for representation learning, and neural architecture search for contrastive learning.

\subsection{Differentiable Neural Architecture Search}
\label{seq:darts}
One of the most trending differentiable NAS family of methods is derived from Differentiable ARchiTecture Search (DARTS) \cite{liu2019darts}. This method uses Stochastic Gradient Descent (SGD) to optimize a set of architectural parameters (denoted $\alpha$) that represent operations inside building blocks (i.e., elementary components of the network) called "cells". A cell $C$ can be considered as a direct acyclic graph whose nodes represent states. Its edges are a mix of $K$ different operations $O = \{o_{1},...,o_K\}$ that define the search space $S$. Multiple cells can be stacked up to form a global "supernet" that encompasses all candidate architectures. Furthermore, as part of a weight-sharing mechanism, DARTS only searches for two types of cells: \textit{normal} cells (i.e., cells that make up most of the network) and \textit{reduction} cells (i.e., cells that perform dimension reduction). The \textit{reduction} cells are typically positioned at the 1/3 and 2/3 of the supernet. As the SGD on $\alpha$ occurs while the supernet containing all cells is being trained on a given dataset, DARTS is practically solving a bi-level optimization problem. Moreover, the $\alpha$ weights are relaxed into a continuous form by discretization through a softmax \cite{softmax1990} operation as follows:
\begin{equation}
    \overline{o}_{i,j}(x) = \sum^{K}_{k=1} \frac{exp(\alpha^{k}_{i,j})}{\sum^{K}_{k'=1} exp(\alpha^{k'}_{i,j})}o_{k}(x)
\label{eq:softmax}
\end{equation}
where $\overline{o}_{i,j}(x)$ is the mixed output of edge $e_{i,j}$ for input feature $x$ and $\alpha^{k}_{i,j} \in \alpha_{i,j}$ is the weight associated with operation $o_{k} \in O$ for $e_{i,j}$.

Several works attempted to improve on DARTS. P-DARTS \cite{chen2021progressive} significantly reduced the search time by progressively deepening the architecture when searching, leading to a better search space approximation and regularization.
PC-DARTS \cite{xu2019pc} attempted to reduce DARTS' memory cost by sampling only a portion of the supernet to avoid redundancy in the search space exploration. FairDARTS \cite{chu2020fair} tried to solve two critical problems that occurred in DARTS, the over-representation of \textit{skip} connections and the uncertainty in the probability distribution of operations. To this end, the authors used the sigmoid function rather than softmax (see Eq. \ref{eq:softmax}) and crafted a novel loss function that can push $\alpha$ values towards 0 or 1. DARTS- \cite{chu2020darts} introduced auxiliary skip connections that are less prone to become dominant, thus ensuring fairer competition with the other operations. $\beta$-DARTS \cite{ye2022beta} introduced a new regularization method, called \textit{Beta-Decay}, that prevents the architectural parameters from saturating. \textit{Beta-Decay} led to increased robustness and better generalization ability. Finally, D-DARTS \cite{heuillet2021d} proposed a mechanism to distribute the search process to the cell level. This approach led to the individualization of each cell, thus increasing the diversity among the candidate architectures and expanding the search space.

\subsection{Siamese Neural Networks}
Bromley et al.~\cite{bromley1993signature} first proposed the Siamese Neural Networks (SNNs) architecture as \say{twin} (i.e., identical and sharing the same weights) models that process two or more inputs and compare their outputs. The central intuition behind this concept is that comparing the output feature vectors will highlight the discrepancies between the inputs. Hence, this approach is advantageous in signature \cite{bromley1993signature} or face \cite{taigman2014deepface} recognition applications.

Another application of SNNs is unsupervised representation learning, also designated as Self-Supervised Learning (SSL). In particular, it is possible to learn representations from unlabeled data by feeding variations of the same input to twin Convolutional Neural Network (CNN) \cite{lecun1995convolutional} models and computing the similarity between the output feature vectors. This similarity metric is used as a loss function, leading the SNNs to learn robust representations (i.e., resisting disturbance in the input data). This process is denoted as contrastive unsupervised learning. Momentum Contrast (MoCo) \cite{he2020momentum} pre-trains a CNN using unsupervised learning with a momentum encoder and fine-tunes its classifier head on standard supervised linear classification. The authors of MoCo show that the unsupervised pre-trained approach can surpass standard CNN on multiple ImageNet \cite{ILSVRC15} computer vision tasks. SimCLR \cite{chen2020simple} added a Multi-Layer Perceptron (MLP) head as a predictor and highlighted the critical role of strong data augmentation and large batches (e.g., ~8000) in contrastive learning. Following up on this, \cite{chen2020improved} proposed an improved version of MoCo (dubbed MoCo V2) that added a two-layer MLP head in the encoder and modified the data transforms according to those of SimCLR. Bootstrap Your Own Latent (BYOL) \cite{grill2020bootstrap} proposed an SNN framework centered around an \textit{online} network and a \textit{target} network. The output of the \textit{target} network is iteratively bootstrapped to serve as input to the \textit{online} network. The authors showed that BYOL could learn more robust representations than previous approaches. Finally, SimSiam \cite{chen2021exploring} introduced a simpler SNN architecture that removes the need for negative sample pairs, momentum encoders, and large batches. More specifically, SimSiam implements a \textit{stop-grad} mechanism that stops gradient backpropagation in one of the two branches of the twin model. Despite being a more straightforward approach than previous baselines, SimSiam reaches a competitive score on ImageNet classification.

\subsection{Neural Architecture Search for Contrastive Self-Supervised Learning}

A handful of previous works have already explored using NAS for contrastive Self-Supervised Learning (SSL). \cite{kaplan2020self} first introduced a method to leverage NAS to improve existing SSL frameworks such as SimCLR \cite{chen2020simple}. Their approach, dubbed SSNAS, is derived from DARTS \cite{liu2019darts} and reached competitive performance compared with supervised models.
\cite{nguyen2021csnas} proposed CSNAS, a novel way to search for SSL-focused CNN architectures using Sequential Model-Based Optimization. CSNAS leverages a cell-based search space similar to DARTS \cite{liu2019darts} and performs contrastive SSL using PIRL \cite{misra2020pretext}. The authors showed that CSNAS managed to overperform or match both handcrafted architectures and supervised NAS models on image classification tasks. Another work of note is SSWP-NAS \cite{li2022towards}, a proxy-free weight-preserving NAS method for SSL. Similarly to CSNAS, SSWP-NAS is based on DARTS and navigates through a cell-based search space to discover new CNN architectures. SSWP-NAS overperformed previous SSL NAS methods and reached competitive results compared to supervised NAS approaches.
In a drastically different approach, Contrastive Neural Architecture Search (CTNAS) \cite{chen2021contrastive} refactors NAS with Contrastive Learning. A Neural Architecture Comparator is designed to drive the search process by comparing candidate architectures with a baseline architecture. Thus, in this approach, contrary to other works, Contrastive Learning is used to enhance NAS rather than the other way around.

In this article, we propose to go further than the previous works listed above by using differentiable NAS to directly enhance the Siamese (i.e., MLP) architecture rather than improve the backbone CNN (which is similar to what trending NAS frameworks such as DARTS \cite{liu2019darts} or FBNet \cite{dai2021fbnetv3} do). In Section \ref{seq:experiments}, we show that our NASiam approach is able to discover novel Siamese architectures reaching higher performance than standard Contrastive SSL frameworks such as SimCLR \cite{chen2020simple} or MoCo \cite{chen2020improved}.  

\section{Proposed Approach}
\label{sec:approach}
This section highlights the key ideas behind our proposed approach: searching for the multi-layer perceptron components of the encoder/predictor pair and crafting an original search space specific to contrastive learning with siamese neural networks.

\subsection{Searching for an Encoder/Predictor Pair}
\label{seq:nasim}
First, we focused on SimSiam \cite{chen2021exploring} as a simple baseline upon which to build our approach.
SimSiam uses a siamese architecture consisting of an encoder $f$ and a predictor $h$. The encoder $f$ is composed of a baseline CNN (e.g., ResNet50 \cite{he2016deep}) and of a projector head (i.e., a three-layer MLP) that is duplicated on twin branches that take variations of the same image as input. A two-layer MLP $h$ is then added on top of one of the branches to act as a predictor head. The discrepancy between the output feature vectors of the two branches is computed using a contrastive loss as follows:
\begin{equation}
    \label{eq:contrastive_loss}
    \mathcal{L} = \frac{1}{2}(\mathcal{D}(p_1,\texttt{stopgrad}(z_2))+\mathcal{D}(p_2,\texttt{stopgrad}(z_1))
\end{equation}
where $z_1 = f(x_1)$, $z_2 = f(x_2)$, $p_1 = h(z_1)$, $p_2 = h(z_2)$ for input images $x_1$ and $x_2$, \texttt{stopgrad} is a mechanism that stops gradient backpropagation (in other words, the argument inside \texttt{stopgrad} is detached from the gradient computation), and $\mathcal{D}$ is the negative cosine similarity defined as follows:
\begin{equation}
    \mathcal{D}(p, z) = - \frac{p}{||p||_2}.\frac{z}{||z||_2}
\end{equation}
where $||.||_2$ is the $l_2$ norm.

In our proposed approach, we kept most of the global structure of the underlying siamese framework. However, we used a Differentiable NAS method to search for an encoder projector head architecture up to $n$ layers and a predictor architecture up to $m$ layers. More specifically, we consider a set $O = \{o_{1},...,o_K\}$ of candidate operations. We search for two cells (see Section \ref{seq:darts}) $C_e$ and $C_p$ for the encoder and decoder respectively. Contrary to DARTS \cite{liu2019darts}, each cell is structured as a linear sequence of layers where each layer is a mixed output of $|O| = K$ operations. Each operation $o$ in each layer $i$ is weighted by a parameter $\alpha_i^{o}$. The sets of architectural parameters for $C_e$ and $C_p$ are denoted $\alpha_e$ and $\alpha_p$ respectively. Similarly to Eq. \ref{eq:softmax}, operation values in each layer are discretized as follows:
\begin{equation}
\label{eq:discretization}
    \overline{o}_{i}(x) = \sum^{K}_{k=1}\sigma_{SM}(\alpha^k_{i})o_{k}(x)
\end{equation}
where $\overline{o}_{i}$ is the mixed operation of layer $i$, $\alpha^k_{i}$ is the architectural weight assigned to $o_{k} \in O$ for layer $i$, and $\sigma_{SM}$ denotes the \textit{softmax} operation.
The supernet encompassing $f$ and $h$ is trained on a portion of a dataset while $C_e$ and $C_p$ are simultaneously searched on another portion of the same dataset. Hence, we solve a bi-level optimization problem formulated as
\begin{equation}
    \begin{split}
        \underset{\alpha_e, \alpha_p}{\text{min }} \mathcal{L}_{val}(w^{*}(\alpha_e, \alpha_p),\alpha_e, \alpha_p),\\
    \text{s.t.} w^{*}(\alpha_e, \alpha_p) = \underset{w}{\text{argmin }} \mathcal{L}_{train}(w,\alpha_e, \alpha_p),
    \end{split}
\end{equation}
 where $w$ denotes the supernet weights, $\mathcal{L}_{train}(w, \alpha_e, \alpha_p) = \mathcal{L}(w, \alpha_e, \alpha_p)$ is the training loss, and $\mathcal{L}_{val}(w^{*}, \alpha_e, \alpha_p) = \mathcal{L}(w^{*}, \alpha_e, \alpha_p)$ is the validation loss. 

Once the search phase is complete, for each layer $i$ of each cell, we select the best-performing operation according to the discretized weights $\alpha_e$ and $\alpha_p$ to form the encoder/predictor architecture genotype $G$.
The whole neural architecture search process is detailed in Algorithm \ref{algo:parsing}.

\begin{algorithm}[ht]
\caption{Algorithm describing the differentiable neural architecture search process of NASiam}
\small
\label{algo:parsing}
\begin{algorithmic}
\REQUIRE Object: $C_e$, encoder cell containing architectural weights
\REQUIRE Object: $C_p$, predictor cell containing architectural weights
\REQUIRE List: $O$, list of operations
\REQUIRE List: $D_t$, train dataset 
\REQUIRE List: $D_v$, validation dataset 
\REQUIRE Object: $model$, backbone CNN model
\REQUIRE Object: $opt$, model optimizer
\REQUIRE Object: $search\_opt$, search optimizer
\REQUIRE Integer: $E$, number of epochs
\FOR{$e$ in [0, $E$[}
    \FOR{$(x_1, x_2)$ in $D_v$}
       \STATE $(x_1, x_2) \gets model(x_1, x_2)$
       \STATE $(z_1, z_2) \gets C_e(x_1, x_2)$
       \STATE $(p_1, p_2) \gets C_p(x_1, x_2)$
       \STATE $\text{stop\_grad}(z_1, z_2)$
       \STATE $loss \gets -0.5(\text{cosine\_similarity}(p_1, z_2) + \text{cosine\_similarity}(p_2, z_1))$
       \STATE $search\_opt.\text{optimization\_step}(loss, C_e.weights, C_p.weights)$
    \ENDFOR
    \FOR{$(x_1, x_2)$ in $D_t$}
       \STATE $(x_1, x_2) \gets model(x_1, x_2)$
       \STATE $(z_1, z_2) \gets C_e(x_1, x_2)$
       \STATE $(p_1, p_2) \gets C_p(x_1, x_2)$
       \STATE $\text{stop\_grad}(z_1, z_2)$
       \STATE $loss \gets -0.5(\text{cosine\_similarity}(p_1, z_2) + \text{cosine\_similarity}(p_2, z_1))$
       \STATE $opt.\text{optimization\_step}(loss, C_e.weights, C_p.weights)$
    \ENDFOR
\ENDFOR
\STATE $A \gets \text{empty\_list()}$
\FOR{$C$ in $\{C_e, C_p\}$}
\STATE $C_{f} \gets \text{empty\_list()}$
\STATE $n \gets |C|$
\FOR{$i$ in $[0, n[$}
    \STATE $op \gets \text{argmax}(C.weights[i])$
    \STATE $\text{append}(op, C_{f})$ 
\ENDFOR
\STATE $\text{append}(C_f, A)$
\ENDFOR
\RETURN $A$
\end{algorithmic}
\end{algorithm}

\begin{figure*}[ht]
    \centering
    \includegraphics[width=\linewidth]{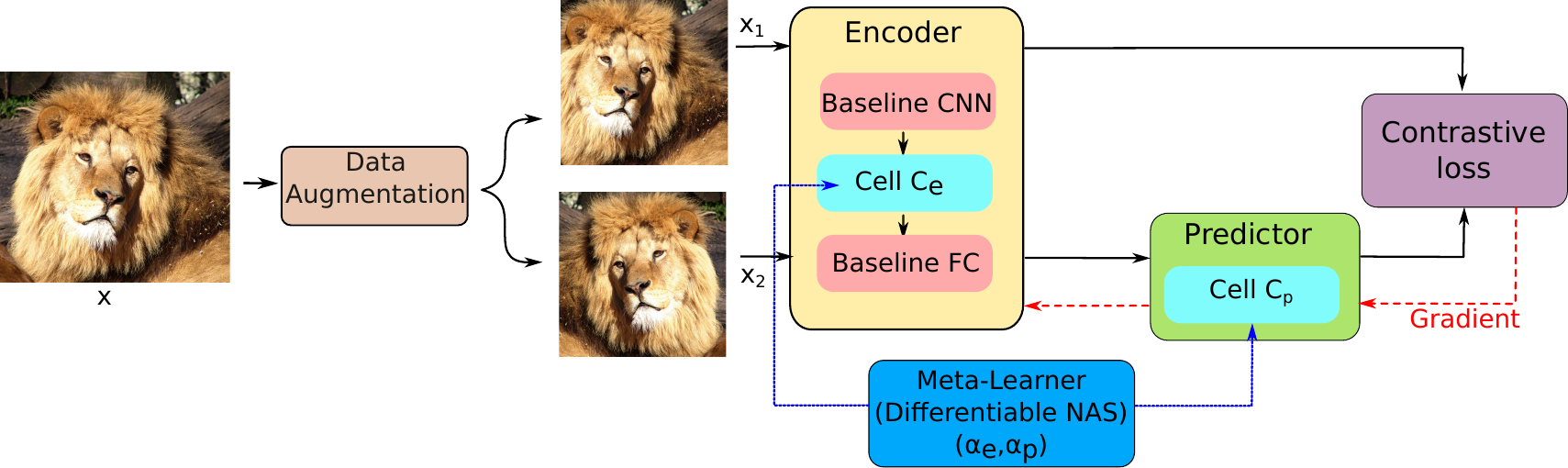}
    \caption{Layout of the NASiam architecture. Siamese network encoder/predictor (projection MLPs) architectures are searched using differentiable NAS wrapped around a Siamese framework such as SimSiam \cite{chen2020simple}, which is the baseline used in the present figure. First, an input image x is augmented to produce two variations $x_1$ and $x_2$. Each of these two inputs is then fed into one of the two branches of the siamese network. While both $x_1$ and $x_2$ go through an encoder equipped with an MLP projection head, $x_2$ is further processed by an MLP predictor. Finally, a negative cosine contrastive loss is computed and backpropagated to minimize the similarity between the two branches' output feature maps. Both the encoder and the decoder contain cells (i.e., $C_e$ and $C_p$, respectively) that are designed using a differentiable NAS approach. Architectural parameters for $C_e$ and $C_p$ are denoted $\alpha_e$ and $\alpha_p$ respectively.} 
    \label{fig:nasim}
\end{figure*}

Our approach, dubbed NASiam (Neural Architecture Search for Siamese Networks), is summarized in Fig. \ref{fig:nasim}. In addition to SimSiam, we also experimented NASiam on other Siamese frameworks such as SimCLR\cite{chen2020simple} and MoCo V2 \cite{chen2020improved}. However, those frameworks (SimCLR and MoCo V2) do not rely on a predictor. Hence, in that case, we only performed NAS for the MLP projector head of the encoder (i.e., only searching for cell $C_e$).

In Section \ref{seq:experiments}, we show that NASiam can consistently improve the performance of popular siamese frameworks (SimSiam, SimCLR, MoCo, and BYOL) in both small-scale (CIFAR-10 and CIFAR-100 \cite{krizhevsky2009learning}) and large-scale (ImageNet \cite{ILSVRC15}) image classification datasets.


\subsection{Crafting a Contrastive Learning-Specific Search Space}
\label{seq:searchspace}
To accompany our novel NASiam approach (see Section \ref{seq:nasim}), we crafted an original search space $S$ specifically designed for MLPs. $S$ comprises the following 7 operation blocks: \texttt{linear} + \texttt{batch\_norm} + \texttt{ReLU}, \texttt{linear} + \texttt{batch\_norm} + \texttt{Hardswish}, \texttt{linear} + \texttt{batch\_norm} + \texttt{SiLU}, \texttt{linear} + \texttt{batch\_norm} + \texttt{ELU}, \texttt{max\_pool\_3x3} (1-dimensional) + \texttt{batch\_norm}, \texttt{avg\_pool\_3x3} (1-dimensional) + \texttt{batch\_norm}, and \texttt{Identity} (\textit{skip connection}). Hence, $S$ includes several types of fully connected layers, each featuring a different activation function. The motivation behind adding activation functions to the search space is to increase diversity among the candidate architectures and explore alternatives to the classic ReLU function (e.g., Hardswish \cite{howard2019searching}, or Mish \cite{misra2020mish}). To that end, it also makes it possible to mix different activation functions according to the type of network (i.e., projector or predictor) and the location inside that network. In contrast, previous baselines \cite{chen2020simple, chen2021exploring, chen2020improved} only relied on a single activation for both networks regardless of their respective architectures.

While unconventional, including pooling layers in the search space is helpful, as we show in Section \ref{seq:experiments} that they can help prevent collapsing. Moreover, the authors of SimSiam \cite{chen2021exploring} indicated that insufficient or too many Batch Normalization (BN) layers could cause the model to underperform severely or become unstable. They empirically demonstrate that the optimal setting for SimSiam is to place BNs after every layer except for the predictor's output layer. Hence, we follow this assertion by adding BNs after every \texttt{linear} and \texttt{pooling} operation except for the predictor's final layer. Finally, we also included the \texttt{identity} operation so that the search algorithm can modulate the number of layers in the architecture. This way, we can indicate a maximum number of layers $n$, and the search algorithm can craft an architecture of size $m < n$ by \say{skipping} layers.

\section{Experiments}
\label{seq:experiments}
This section presents the results of our image classification experiments on small-scale (CIFAR-10, CIFAR-100) and large-scale (ImageNet) datasets.

\subsection{Experimental Settings}
\label{sec:settings}
We used RTX 3090 and Tesla V100 Nvidia GPUs to conduct our experiments. We searched for predictor/encoder pairs for 100 epochs on CIFAR-10, and CIFAR-100 \cite{krizhevsky2009learning} using the \texttt{SGD} optimizer with $lr = 0.06$, $wd = 5e-4$, and a batch size of 512. We set a maximum of 6 layers for the encoder. If the baseline siamese framework relies on a predictor, we search for a 4-layer predictor architecture. The whole search process on these settings takes around 2.3 GPU hours on a single GPU. We did not search on the full ImageNet \cite{ILSVRC15} dataset as it is prohibitively expensive (i.e., it takes around 12 GPU days on a single GPU). Instead, we transferred our best CIFAR-searched architecture to ImageNet. For the pre-training and linear classification phases, we kept the same settings as \cite{chen2021exploring}. Our code is based on PyTorch 1.12.

\subsection{Ablation Study on the Importance of Pooling Layers}
\label{sec:ablation}
\begin{figure}[ht]
    \centering
    \includegraphics[width=0.8\linewidth]{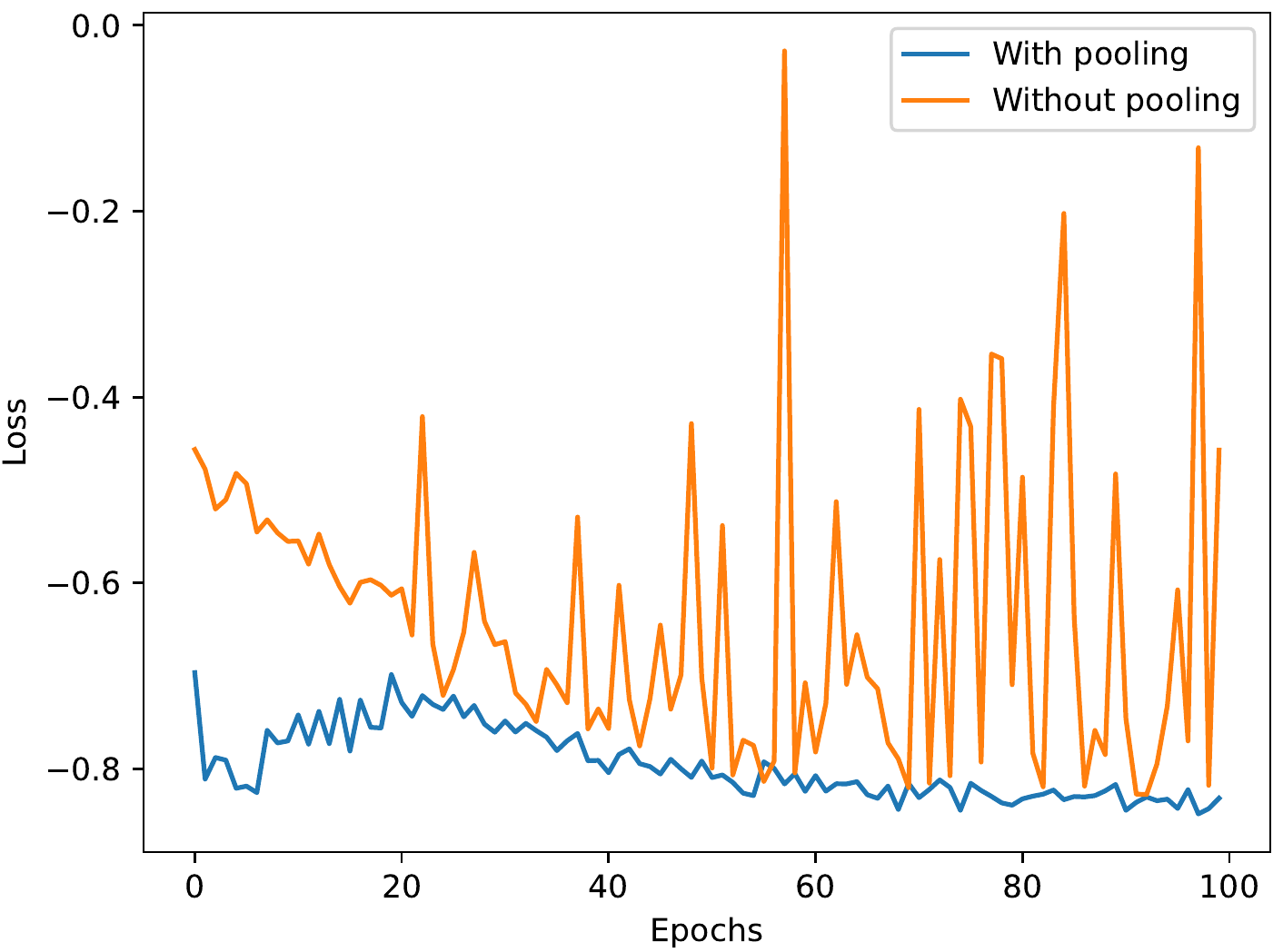}
    \caption{Plot of the negative cosine contrastive loss while pretraining two NASiam models on CIFAR-10. The baseline framework is SimSiam with a ResNet18 backbone. The two models are searched on search spaces $S$ (blue line) and $S'$ (red line) respectively. The model searched on $S$ achieves better similarity, thus making the relevance of pooling layers clear.}
    \label{fig:pooling_comparison}
\end{figure}

We conducted an ablation study on the importance of including pooling layers in our novel space search $S$ (see Section \ref{seq:searchspace}). To this end, we simply removed \texttt{max\_pool\_3x3} and \texttt{avg\_pool\_3x3} from $S$ to form $S'$. When comparing the results in Table \ref{tab:tab_pool}, we can observe that, when searching on $S'$ rather than $S$, the validation top-1 accuracy of NASiam drops significantly (by around 3 \%). Moreover, when analyzing the genotypes searched on CIFAR-10 and CIFAR-100 using $S$, it appears that the predictor architectures always contain pooling layers (making up to 40 \% of the total architecture). In addition, Fig. \ref{fig:pooling_comparison} shows that the model (see Eq. \ref{eq:contrastive_loss}) achieved better similarity and faster convergence when searched on $S$ rather than $S'$. Thus, these findings highlight the critical role pooling layers play in ensuring high performance and preventing collapse, especially concerning the encoder architecture.
\begin{table}[ht]
    \caption{Results on CIFAR-10 linear classification of two NASiam models using search space $S$ and $S'$ respectively. Both models were pre-trained for 100 epochs.  The baseline framework is SimSiam with a ResNet18 backbone.}
    \label{tab:tab_pool}
    \begin{center}
    \footnotesize
    \begin{tabular}{lllll}
        \toprule
        \thead{Search\\ Space} & \thead{Search\\ Epochs} & \thead{Pre-Train\\ Epochs}  & \thead{ Validation \\ Top-1 (\%)} & \thead{Validation\\ Top-5}\\
        \midrule
        $S$ & 50 & 100 & \textbf{66.6} & 91.3\\
        $S'$ & 50 & 100 & 63.7 & 86.1 \\
        \bottomrule
    \end{tabular}
    \end{center}
\end{table}

\subsection{Incidence of Data Augmentations on the NAS process}

In self-supervised learning, data augmentations are paramount to prevent the model from overfitting and the contrastive loss (see Eq. \ref{eq:contrastive_loss}) from saturating to -1. In contrast, differentiable neural architecture search methods \cite{liu2019darts, chu2020fair, heuillet2021d} scarcely employ data augmentation as they only train the supernet for a small number of epochs (e.g., 50). Thus, a legitimate interrogation is how the strong data augmentation policy used in SSL frameworks can interfere with the differentiable search process. To answer this question, we searched for two different SimSiam \cite{chen2021exploring} models with and without the data augmentation policy activated on CIFAR-10 and compared the resulting architectures. 

Table \ref{tab:tab_data_aug} shows that deactivating the data augmentation policy leads to a degenerated architecture with a dominance of \texttt{skip connections} (50 \% of the architecture) associated with performance collapse. Furthermore, Fig. shows that, during the search phase, the similarity loss converges significantly faster towards -1, thus presenting a collapsing behavior. This observation correlates with the architectural collapse described in numerous differentiable NAS studies \cite{chu2020fair, Zela2020Understanding, ye2022beta}. This collapsing behavior is akin to overfitting for NAS and is caused by the high prominence of \textit{skip connections} due to their unfair advantage (compared to parametric operations). Thus, data augmentation clearly has a positive impact on the differentiable search process and should not be deactivated, in contrast with supervised learning.

\begin{table}[ht]
    \caption{Results on CIFAR-10 linear classification of two NASiam models using either SimSiam data augmentation policy or no data augmentation. Both models were pre-trained for 800 epochs.  The baseline framework is SimSiam with a ResNet18 backbone.}
    \label{tab:tab_data_aug}
    \begin{center}
    \footnotesize
    \begin{tabular}{lllll}
        \toprule
        \thead{Data\\ Augmentation} & \thead{Search\\ Epochs} & \thead{Pre-Train\\ Epochs}  & \thead{ Validation \\ Top-1 (\%)} & \thead{\% of \\ Skip Connections}\\
        \midrule
        Yes & 100 & 800 & 91.2 & 20\\
        No & 100 & 800 & 10.0 & 50\\
        \bottomrule
    \end{tabular}
    \end{center}
\end{table}

\begin{figure}
    \centering
    \includegraphics[width=0.8\linewidth]{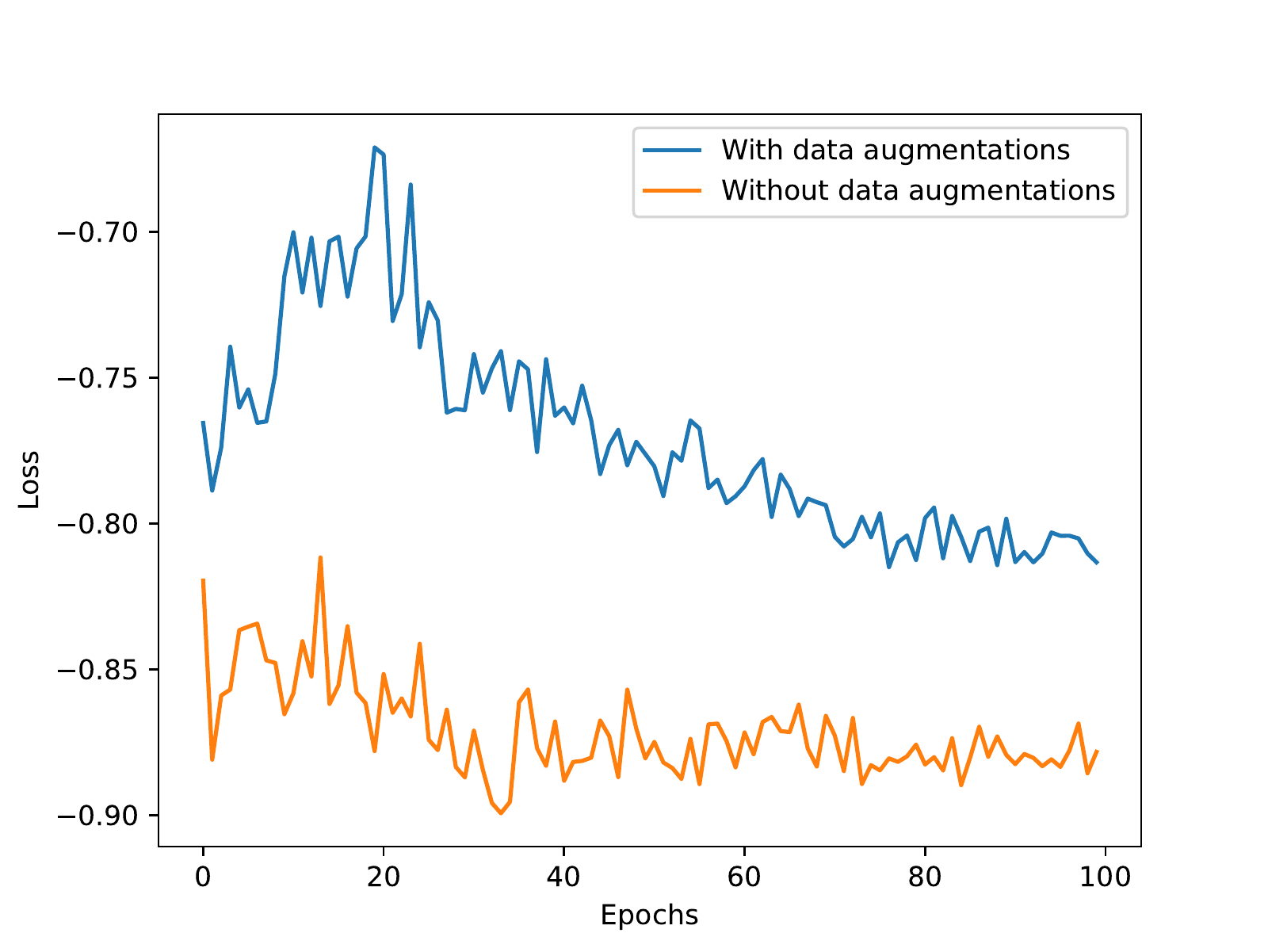}
    \caption{Plot of the negative cosine contrastive loss while pretraining two NASiam models on CIFAR-10. The baseline framework is SimSiam with a ResNet18 backbone. The two models are searched with and without data augmentation respectively.}
    \label{fig:graph_data_aug}
\end{figure}

\subsection{Preliminary Results on CIFAR}
\label{sec:results_cifar}

\begin{figure}[ht]
    \centering
    \includegraphics[width=0.8\linewidth]{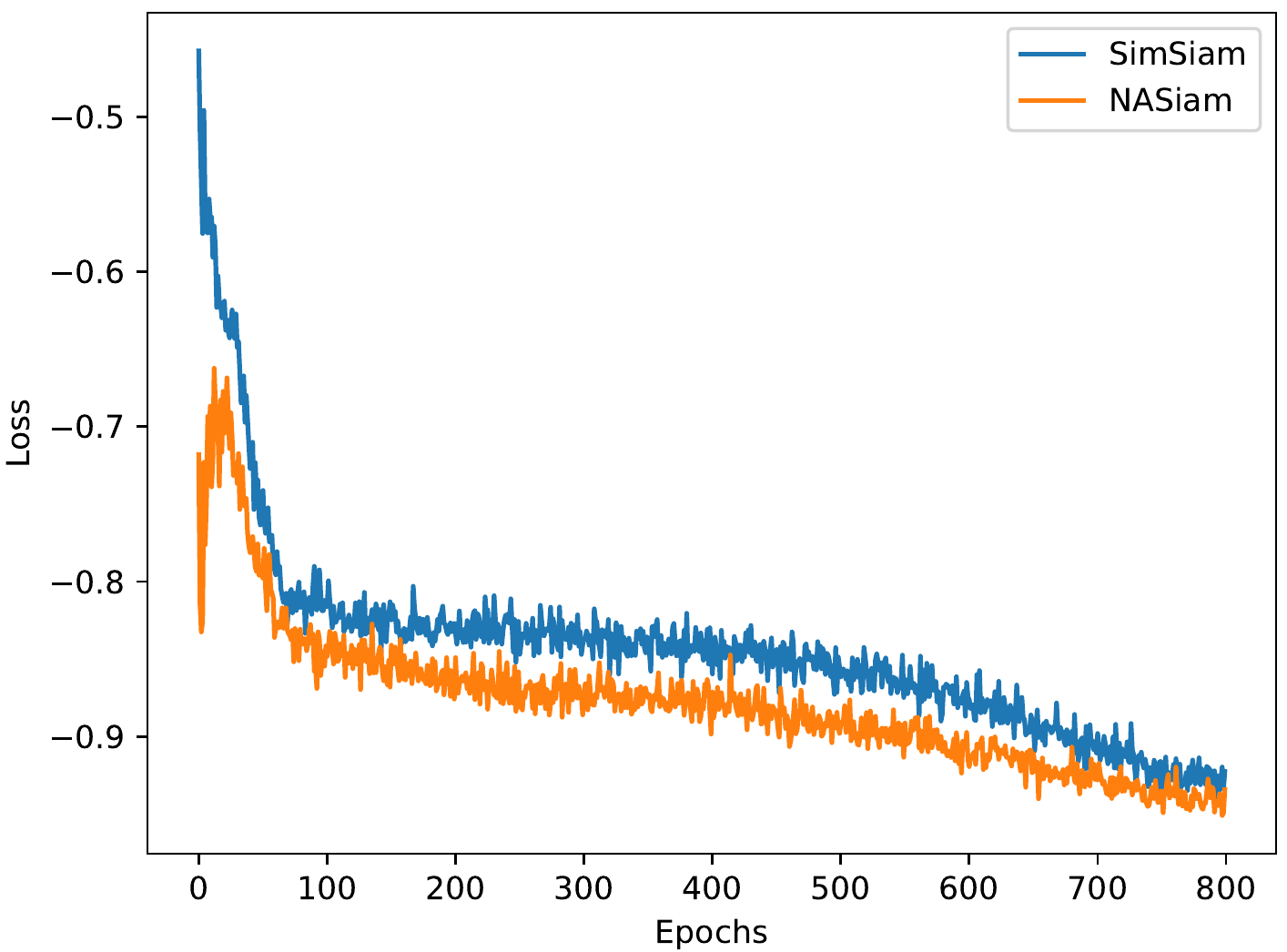}
    \caption{Plot of the negative cosine contrastive loss when pretraining SimSiam and NASiam for 800 epochs on CIFAR-10. NASiam converges faster without collapsing and achieves better similarity than SimSiam.}
    \label{fig:pretrain}
\end{figure}

To quickly assess the behavior of our novel approach NASiam, we first conducted preliminary experiments on small-scale CIFAR datasets \cite{krizhevsky2009learning}. We searched NASiam architectures for 100 epochs on CIFAR-10 and CIFAR-100 using the CIFAR version of ResNet18 \cite{he2016deep} as the encoder backbone. Then, we performed unsupervised pretraining for 800 epochs with a cosine annealing schedule before training a linear classifier using frozen features for 100 epochs. In these settings, NASiam overperforms SimSiam by 1.4 \% and 0.4 \% on CIFAR-10 and CIFAR-100 respectively (see Table \ref{tab:cifar10} and Table \ref{tab:cifar100}). In addition, Fig. \ref{fig:pretrain} shows us that NASiam can achieve better similarity than SimSiam without saturating the contrastive loss to $-1$ (i.e., a \say{collapsing} behavior). Furthermore, results were also positive when using alternative siamese frameworks, with NASiam overperforming both MoCo V2 \cite{chen2020improved} and SimCLR \cite{chen2020simple}.

\begin{table}[ht]
    \caption{Results of pre-training for 800 epochs on CIFAR-10 linear classification with SGD. The backbone is the CIFAR version of ResNet18. $\dagger$: Result obtained by running the official implementation with the hyperparameters suggested by the authors for CIFAR-10.}
    \label{tab:cifar10}
    \footnotesize
    \begin{center}
    \begin{tabular}{lllll}
        \toprule
        \thead{Model} & \thead{Batch\\ Size} & \thead{Pre-Train\\ Epochs} & \thead{Train\\ Epochs}  & \thead{ Validation \\ Top-1 (\%)}\\
        \midrule
         MoCo V2 \cite{chen2020improved}$\dagger$ & 256 & 800 & 100 & 89.8\\
        SimCLR \cite{chen2020simple}$\dagger$ & 256 & 800 & 100 & 91.1\\
        SimSiam \cite{chen2021exploring}$\dagger$ & 256 & 800 & 100 & 89.5\\
        \multirow{3}{*}{Ours} \hspace{0.8em} NASiam (SimSiam) & 256 & 800 & 100 & \textbf{91.2}\\
        \hspace{3em} NASiam (MoCo V2) & 256  & 800 & 100 & \textbf{90.4}\\
        \hspace{3em} NASiam (SimCLR) & 256  & 800 & 100 & \textbf{92.1}\\
        \bottomrule
    \end{tabular}
    \end{center}
\end{table}

\begin{table}[ht]
    \caption{Results of training for 800 epochs on CIFAR-100 linear classification with SGD. The backbone is the CIFAR version of ResNet18. $\dagger$: Result obtained by running the official implementation with the hyperparameters suggested by the authors for CIFAR-10.}
    \label{tab:cifar100}
    \footnotesize
    \begin{center}
    \begin{tabular}{lllll}
        \toprule
        \thead{Model} & \thead{Batch\\ Size} & \thead{Pre-Train\\ Epochs} & \thead{Train\\ Epochs}  & \thead{ Validation \\ Top-1 (\%)} \\
        \midrule
        MoCo V2 \cite{chen2020improved}$\dagger$ & 256 & 800 & 100 & 62.9\\ 
        SimCLR \cite{chen2020simple}$\dagger$ & 256 & 800 & 100 & 63.6\\
        SimSiam \cite{chen2021exploring}$\dagger$ & 256 & 800 & 100 & 63.7\\
        \multirow{3}{*}{Ours} \hspace{0.8em} NASiam (SimSiam) & 256 & 800 & 100 & \textbf{64.1}\\
        \hspace{3em} NASiam (MoCo V2) & 256 & 800 & 100 & \textbf{65.0}\\
        \hspace{3em} NASiam (SimCLR) & 256 & 800 & 100 & \textbf{68.9}\\
        \bottomrule
    \end{tabular}
    \end{center}
\end{table}

\subsection{Results on ImageNet}
\label{sec:results_imagenet}

We conducted image classification experiments on ImageNet \cite{ILSVRC15} as a standard practice to evaluate the performance of our novel approach on large-scale datasets. As stated in Section \ref{sec:settings}, we transferred our best CIFAR architecture instead of searching directly on ImageNet to save computational resources. Then, we performed unsupervised pretraining on ImageNet for 100 epochs before training a linear classifier with frozen features for 100 epochs. The results are presented in detail in Table \ref{tab:imagenet}. As for CIFAR (see \ref{sec:results_cifar}), NASiam consistently achieves better linear classification results than the baseline frameworks, thus validating its usefulness.

\begin{table}[ht]
    \caption{Results of training for 100 epochs on ImageNet linear classification with SGD. The backbone is ResNet50. Models were pre-trained for 100 epochs on ImageNet. $\dagger$: Score obtained by the authors of SimSiam by using an improved version of the model.}
    \label{tab:imagenet}
    \footnotesize
    \begin{center}
    \begin{tabular}{lllll}
        \toprule
        \thead{Model} & \thead{Batch\\ Size} & \thead{Pre-Train\\ Epochs} & \thead{Train\\ Epochs}  & \thead{ Validation \\ Top-1 (\%)} \\
        \midrule
        MoCo V2 \cite{chen2020improved}$\dagger$ & 256 & 100 & 100 & 67.4\\ 
        BYOL \cite{grill2020bootstrap}$\dagger$ & 4096 & 100 & 100 & 66.5\\
        SimCLR \cite{chen2020simple}$\dagger$ & 4096 & 100 & 100 & 66.5\\
        SimSiam \cite{chen2021exploring} & 256 & 100 & 100 & 67.1\\
        \multirow{3}{*}{Ours} \hspace{0.8em} NASiam (SimSiam) & 256 & 100 & 100 & \textbf{67.4}\\
        \hspace{3em} NASiam (MoCo V2) & 256 & 100 & 100 & \textbf{67.4}\\
        \hspace{3em} NASiam (SimCLR) & 4096 & 100 & 100 & \textbf{67.2}\\
        \bottomrule
    \end{tabular}
    \end{center}
\end{table}

\subsection{Object Detection and Instance Segmentation Results on COCO}

Table \ref{tab:coco} displays the results of transferring our NASiam models pretrained on ImageNet \cite{ILSVRC15} to Microsoft COCO \cite{lin2014microsoft} object detection and instance segmentation tasks. We can see that NASiam consistently overperforms handcrafted SSL architectures in both tasks. Hence, NASiam architectures can successfully generalize to computer vision tasks other than image classification.

\begin{table*}[ht]
    \begin{center}
    \caption{\label{tab:coco}Comparison of backbone models for MaskRCNN \cite{he2017mask} on COCO \cite{lin2014microsoft} using a 1x schedule and ResNet50 \cite{he2016deep} as the baseline CNN. All models are pretrained for 200 epochs on ImageNet, finetuned for 12 epochs on COCO 2017 train set, and evaluated on COCO 2017 val set.}
    \footnotesize
    \begin{tabular}{lllllll}
        \toprule
        \thead{Models} & \thead{$AP_{50}$} (\%) & \thead{$AP$} (\%) & \thead{$AP_{75}$} (\%) & \thead{$AP^{mask}_{50}$} (\%) & \thead{$AP^{mask}$} (\%) & \thead{$AP^{mask}_{75}$} (\%) \\
        \midrule
        ImageNet supervised & 58.2 & 38.2 & 41.2 & 54.7 & 33.3 & 35.2\\
        SimCLR & 57.7 & 37.9 & 40.9 & 54.6 & 33.3 & 35.3\\
        SimSiam & 57.5 & 37.9 & 40.9 & 54.2 & 33.2 & 35.2\\
        MoCo V2 & 58.8 & 39.2 & 42.5 & 55.5 & 34.3 & 36.6\\
        BYOL & 57.8 & 37.9 & 40.9 & 54.3 & 33.2 & 35.0\\
        \textbf{NASiam (SimSiam)} & \textbf{58.6} & \textbf{39.0} & \textbf{42.1} & \textbf{55.2} &\textbf{34.1} & \textbf{36.3}\\
        \bottomrule
    \end{tabular}
    \end{center}
\end{table*}

\subsection{Discussion on the Composition of the Architectures}

Some facts are noteworthy when comparing encoder/predictor architectures discovered on CIFAR-10 by our novel approach (see Section \ref{seq:nasim}) with those of SimSiam \cite{chen2021exploring}. 

\begin{figure}[ht]
    \centering
    \includegraphics[width=\linewidth]{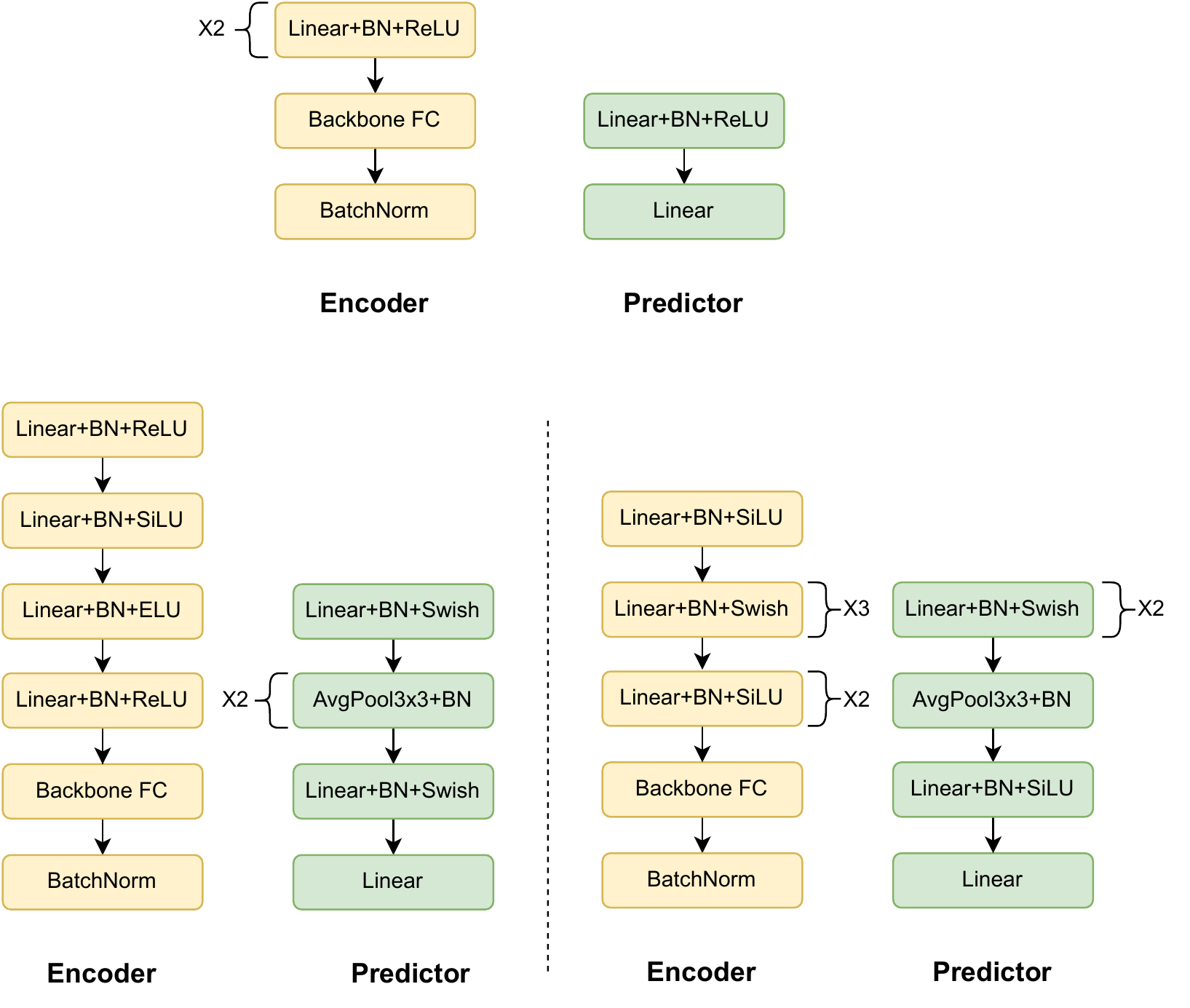}
    \caption{Composition of encoder/predictor pair architectures. (\textbf{Top}) SimSiam model. (\textbf{Bottom left}) NASiam model searched for 100 epochs on CIFAR-10 using SimSiam as the baseline framework with ResNet18 as the backbone CNN. (\textbf{Bottom right}) NASiam model searched for 100 epochs on CIFAR-10 using SimSiam as the baseline framework with ResNet50 as the backbone CNN. ResNet18-searched and ResNet50-searched architectures are clearly different, with ResNet50 needing a deeper encoder.}
    \label{fig:cifar_archs}
\end{figure}

First, in Fig. \ref{fig:cifar_archs}, we can see that both ResNet50 and ResNet18\cite{krizhevsky2009learning} NASiam architectures are significantly deeper than the original SimSiam architecture. Furthermore, a remarkable fact is that the \texttt{ReLU} activation function is in minority in the discovered architectures (and even disappeared completely from the ResNet50 one). Instead, a mix of different activation functions is preferred, with \texttt{SiLU} and \texttt{Hardswish} having a high prominence. Thus, this may indicate that \texttt{ReLU}, despite its popularity, is not the optimal activation function for performing contrastive learning. In addition, the optimizer always selected at least one \texttt{AvgPool3x3+BN} layer to be part of the predictor architecture, hence validating the relevance of including pooling layers in the search space (as already highlighted in Section \ref{sec:ablation}). 

\begin{figure}[ht]
    \centering
    \includegraphics[width=0.8\linewidth]{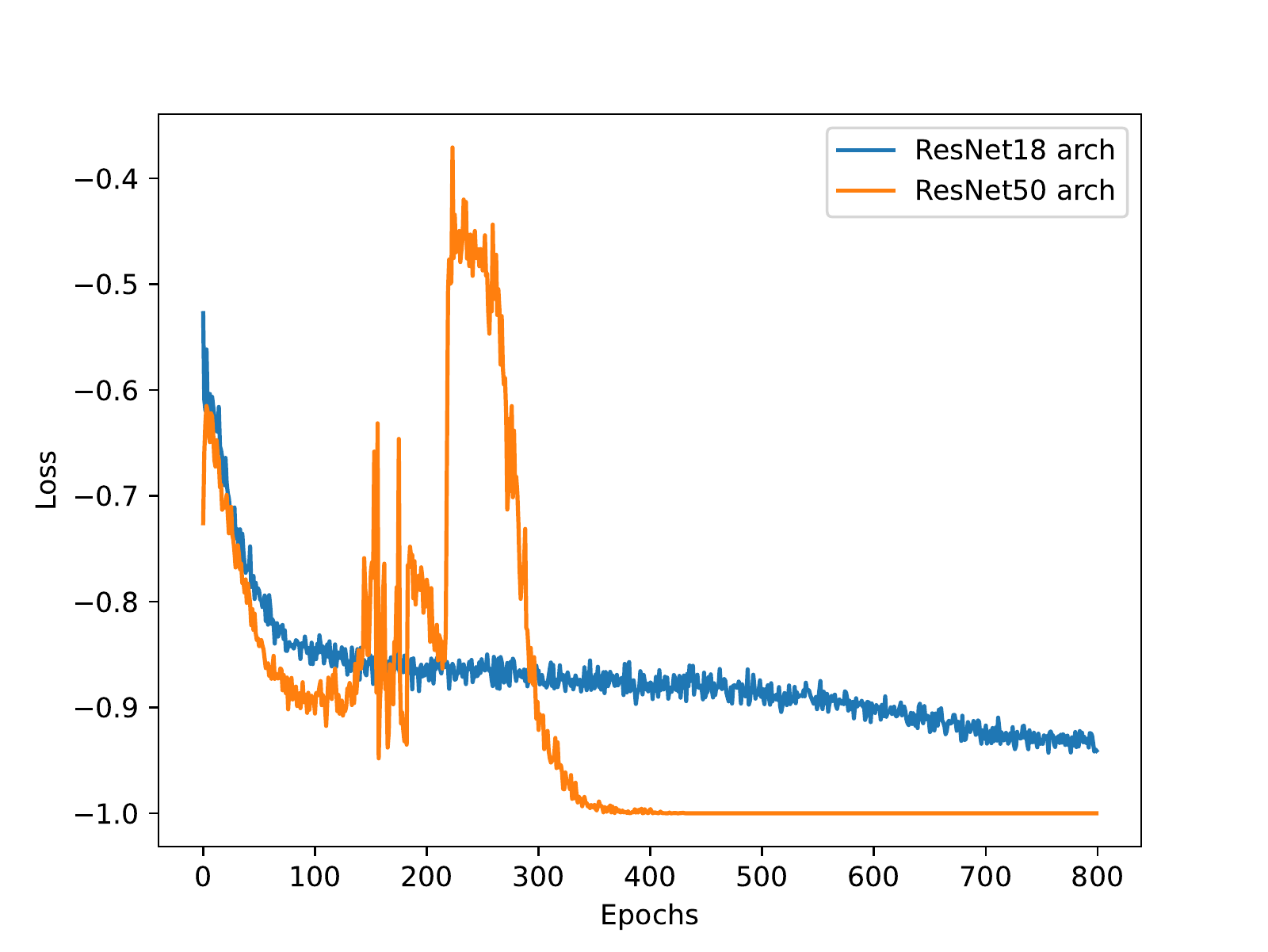}
    \caption{Plot of the negative cosine similarity loss while pretraining NASiam with ResNet18 using architectures searched either with ResNet18 or ResNet50 as backbone. The ResNet50-searched architecture quickly collapses towards -1 and has high variance while the ResNet18-searched one converges as expected.}
    \label{fig:resnet18_vs_resnet50}
\end{figure}

Finally, when comparing both NAS-discovered architectures, we can observe that the ResNet50 one possesses a deeper encoder than the ResNet18-based architecture (i.e., 6 vs. 4 layers), with additional \texttt{Linear+BN+Swish} and \texttt{Linear+BN+SiLU} blocks. However, the two predictor architectures retain the same depth and a similar composition. This is coherent with the recommendations of the authors of SimSiam \cite{chen2021exploring}, where they selected a shallower architecture when training on CIFAR-10 with ResNet18 rather than ResNet50. One hypothesis to explain this discrepancy in architectural sparsity is that ResNet18, being a shallower model than ResNet50, has a less powerful innate ability to extract representations and hence produces less complex feature maps that would not require a deep projector to be analyzed. Using a deeper architecture could even lead to adverse effects. To confirm this hypothesis, we tried to fit a ResNet18 model on CIFAR-10 with the deeper encoder/predictor pair discovered for ResNet50. Fig. \ref{fig:resnet18_vs_resnet50} clearly shows that this architectural setting quickly led to a collapsing behavior (with the contrastive loss rapidly saturating to -1 as soon as epoch 350) with a higher variance than the ResNet18-searched architecture. Hence, this validates the ability of our NASiam approach to discover backbone-specific architectures.

\section{Conclusion}
\label{sec:conclusion}

In this article, we presented NASiam, a novel approach for contrastive learning with siamese networks that searches for efficient encoder/predictor pairs using differentiable neural architecture search (see Section \ref{seq:nasim}). This universal method can enhance many existing siamese frameworks while preserving their underlying structure. In addition, NASiam is efficient as it only costs a few GPU hours. Section \ref{seq:experiments} showed that NASiam discovers encoder/predictor pair architectures that efficiently learn robust representations and overperform previous baselines in small-scale and large-scale image classification datasets. These empirical results support our intuition that the encoder and predictor architectural designs play a decisive role in representation learning. We hope this work will pave the way to further improvements for MLP-headed siamese networks.

\bibliography{refs}
\bibliographystyle{ieeetr}

\end{document}